# Maximum Margin Output Coding

**Yi Zhang**                                              YIZHANG1@CS.CMU.EDU
Machine Learning Department, Carnegie Mellon University

**Jeff Schneider**                                    SCHNEIDE@CS.CMU.EDU
The Robotics Institute, Carnegie Mellon University

## Abstract

In this paper we study output coding for multi-label prediction. For a multi-label output coding to be *discriminative*, it is important that codewords for different label vectors are significantly different from each other. In the meantime, unlike in traditional coding theory, codewords in output coding are to be predicted from the input, so it is also critical to have a *predictable* label encoding.

To find output codes that are both *discriminative* and *predictable*, we first propose a max-margin formulation that naturally captures these two properties. We then convert it to a metric learning formulation, but with an exponentially large number of constraints as commonly encountered in structured prediction problems. Without a label structure for tractable inference, we use overgenerating (i.e., relaxation) techniques combined with the cutting plane method for optimization.

In our empirical study, the proposed output coding scheme outperforms a variety of existing multi-label prediction methods for image, text and music classification.

## 1. Introduction

In traditional channel coding (Cover & Thomas, 1991; Costello & Forney, 2007), a message is encoded into an alternative (and usually redundant) representation so that it can be recovered accurately after being transmitted through a noisy channel. Error-correcting output coding (ECOC) applies the idea of channel coding to multi-class classification (Dietterich & Bakiri, 1995; Allwein et al., 2001) and more recently to multi-label prediction (Hsu et al., 2009; Tai & Lin, 2010; Zhang & Schneider, 2011): we encode the output into a codeword, learn models to predict the codeword, and then recover the correct output from noisy predictions.

In this paper, we study output coding for multi-label prediction and focus on two important issues. First, the coding needs to be **discriminative**: encodings for different outputs should be significantly different from each other, so that the codeword for the correct output will not be confused with incorrect ones, even under noisy predictions. This corresponds to the concept of code distance in coding theory and is related to good error-correcting capabilities (Cover & Thomas, 1991).

Second, output codes should be **predictable**. In output coding, codewords need to be predicted from the input (instead of being actually transmitted through a channel), so it is critical that codewords are easy to predict. From the channel coding perspective, having predictable codewords (and thus low prediction errors) corresponds to reducing the channel error. In multi-label prediction, finding predictable codewords provides an opportunity to exploit the dependency structure in the label space (Zhang & Schneider, 2011).

To design output codes that are both discriminative and predictable, we propose a max-margin formulation defined on the encoding transform. For each sample, the prediction from the input should be close to the encoding of the correct output, and at the same time, the prediction should also be far away from the encoding of any incorrect output. This is naturally captured by maximizing the margin between the prediction distance to correct and incorrect encodings.

We then convert this formulation to a metric learning problem of finding the optimal distance metric in the label space, but with an exponentially large number of constraints as commonly encountered in structured prediction problems. In multi-label prediction, howev-er, the output space does not provide a structure for tractable inference, and we use overgenerating (i.e., relaxation) techniques combined with the cutting plane method to optimize the metric learning formulation. The encoding and decoding operations can be derived from the optimal distance metric in the label space.

We conduct our experiments on multi-label classification of images, text and music. Empirical results show that the proposed output coding scheme outperforms a variety of recent multi-label prediction methods.

## 2. Multi-Label Output Codes: Framework and Existing Methods

In this section, we introduce the general framework for multi-label output coding. Then we review three recently-proposed output codes (Hsu et al., 2009; Tai & Lin, 2010; Zhang & Schneider, 2011), where the encoding is based on random projections, principal component analysis and canonical correlation analysis, respectively. We also argue that these existing output coding schemes are not designed to optimize both discriminability and predictability of the codewords.

### 2.1. Framework

An output coding scheme usually contains three parts: encoding, prediction and decoding. Consider a set of $p$ input variables $\mathbf{x} \in \mathcal{X} \subseteq R^p$ and a set of $q$ output variables $\mathbf{y} \in \mathcal{Y} \subseteq R^q$. In multi-label classification, $\mathbf{y}$ will denote the label vector, and thus $\mathbf{y} \in \mathcal{Y} = \{0,1\}^q$. We have a set of $n$ training examples: $\mathbf{D} = (\mathbf{X}, \mathbf{Y}) = \{(\mathbf{x}^{(i)}, \mathbf{y}^{(i)})\}_{i=1}^n$, where $\mathbf{X}$ and $\mathbf{Y}$ are matrices of size $n \times p$ and $n \times q$, respectively.

**Encoding.** Following previous work (Hsu et al., 2009; Tai & Lin, 2010; Zhang & Schneider, 2011), we consider linear encoding. In this case, the encoding transform can be defined by a set of $d$ linear projections

$$\mathbf{V} = (\mathbf{v}_1, \mathbf{v}_2 \ldots, \mathbf{v}_d) \quad (1)$$

where $d$ is the number of projections, each $\mathbf{v}_k$ ($k = 1, 2, \ldots, d$) is a $q \times 1$ vector representing a projection direction in the label space, and $\mathbf{V}$ is a $q \times d$ matrix.

Given the projection vectors $\mathbf{V}$, the codeword $\mathbf{z}$ for an example $(\mathbf{x}, \mathbf{y})$ is defined:

$$\mathbf{z} = \mathbf{V}^T \mathbf{y} = (\mathbf{v}_1^T \mathbf{y}, \ldots, \mathbf{v}_d^T \mathbf{y})^T \quad (2)$$

where $\mathbf{z}$ is a $d \times 1$ vector. Alternatively, we can also include the $q$ original labels $\mathbf{y} = (y_1, \ldots, y_q)$ into the codeword $\mathbf{z}$, and in this case we have:

$$\mathbf{z} = [\mathbf{I}_q, \mathbf{V}]^T \mathbf{y} = (y_1, \ldots, y_q, \mathbf{v}_1^T \mathbf{y}, \ldots, \mathbf{v}_d^T \mathbf{y})^T \quad (3)$$

where $\mathbf{I}_q$ is a $q \times q$ identity matrix, $\mathbf{V}$ is a $q \times d$ matrix, and $\mathbf{z}$ is a $(q+d) \times 1$ vector.

**Prediction.** After defining the encoding projections $\{\mathbf{v}_k\}_{k=1}^d$, we then learn prediction models from training samples $\{(\mathbf{x}^{(i)}, \mathbf{y}^{(i)})\}_{i=1}^n$ to predict the codeword. For a label projection $\mathbf{v}_k^T \mathbf{y}$ in the codeword, a regression model is usually considered:

$$\hat{m}_k \leftarrow \text{learn\_regression}(\{(\mathbf{x}^{(i)}, \mathbf{v}_k^T \mathbf{y}^{(i)})\}_{i=1}^n) \quad (4)$$

and for an original label $y_j$ ($j = 1, 2, \ldots, q$), a classifier can be learned from training samples:

$$\hat{p}_j \leftarrow \text{learn\_classifier}(\{(\mathbf{x}^{(i)}, y_j^{(i)})\}_{i=1}^n) \quad (5)$$

Given a new sample $\mathbf{x}$, a regression model $\hat{m}_k$ predicts:

$$\hat{m}_k(\mathbf{x}) = E(\mathbf{v}_k^T \mathbf{y} | \mathbf{x}) \quad (6)$$

and a classifier $\hat{p}_j$ predicts:

$$\hat{p}_j(\mathbf{x}) = (\hat{p}_{j0}(\mathbf{x}), \ \hat{p}_{j1}(\mathbf{x})) \quad (7)$$

where

$$\hat{p}_{j0}(\mathbf{x}) = P(y_j = 0 | \mathbf{x}) \quad (8)$$
$$\hat{p}_{j1}(\mathbf{x}) = P(y_j = 1 | \mathbf{x}) \quad (9)$$

**Decoding.** Given a new testing sample $\mathbf{x}$, the decoding procedure recovers the unknown label vector $\mathbf{y}$ from our prediction for the codeword $\mathbf{z}$. The prediction contains $\{\hat{m}_k(\mathbf{x})\}_{k=1}^d$ and optionally $\{\hat{p}_j(\mathbf{x})\}_{j=1}^q$:

$$\hat{\mathbf{y}} \leftarrow \text{decoding}(\mathbf{x}, \{\mathbf{v}_k\}_{k=1}^d, \{\hat{m}_k(\mathbf{x})\}_{k=1}^d, \{\hat{p}_j(\mathbf{x})\}_{j=1}^q) \quad (10)$$

The decoding is usually achieved by maximizing a probability function or minimizing a loss function defined on possible label vector $\mathbf{y}$. Since $\mathbf{y} \in \mathcal{Y} = \{0,1\}^q$, this optimization is usually combinatorial in nature and intractable. As a result, certain approximation is required to obtain the solution $\hat{\mathbf{y}}$, e.g., relaxing $\mathbf{y}$ into a continuous domain and then rounding the relaxed solution (Hsu et al., 2009; Tai & Lin, 2010) or using approximate inference (Zhang & Schneider, 2011).

### 2.2. Coding with compressed sensing

Multi-label compressed sensing (Hsu et al., 2009) is one of the earliest works that formally defines a multi-label output code. For encoding, each projection vector $\mathbf{v}_k \in R^q$ (k = 1, 2, ..., d) is randomly generated as in compressed sensing (Donoho, 2006; Candes, 2006), e.g., a vector with i.i.d. Gaussian or Bernoulli entries. Thus, the codeword $\mathbf{z} = (\mathbf{v}_1^T \mathbf{y}, \ldots, \mathbf{v}_d^T \mathbf{y})^T$ contains random projections of the label vector $\mathbf{y}$.

Decoding follows the sparse approximation algorithms in compressed sensing. Two popular classes are convex relaxation such as $\ell 1$ penalized least squares (Tropp, 2006), and iterative greedy algorithms such as CoSaMP (Needell & Tropp, 2008). For example, an $\ell 1$ penalized convex relaxation solves the following problem:

$$\hat{\mathbf{y}} \leftarrow \underset{\mathbf{y} \in R^q}{\operatorname{argmin}} \ \frac{1}{2} \sum_{k=1}^{d} (\mathbf{v}_k^T \mathbf{y} - \hat{m}_k(\mathbf{x}))^2 + \lambda \sum_{j=1}^{q} |y_j| \quad (11)$$

where $\{\hat{m}_k(\mathbf{x})\}_{k=1}^{d}$ are predictions for the codeword $\mathbf{z} = (\mathbf{v}_1^T \mathbf{y}, \ldots, \mathbf{v}_d^T \mathbf{y})^T$, and the $\ell 1$ penalty $\sum_{j=1}^{q} |y_j| = ||\mathbf{y}||_1$ promotes the sparsity of the solution. Note that this problem is solved in relaxed space $\mathbf{y} \in R^q$.

Use of random projections is justified in compressed sensing, e.g., by the restricted isometry property, that if the true signal $\mathbf{y}$ is sufficiently sparse, one can recover $\mathbf{y}$ from only a small number of random projections. However, from the output coding perspective, random projections do not specifically promote either discriminative or predictable codewords, and thus may not be the most effective method of output coding.

### 2.3. Coding with principal component analysis

Given the $n$ training examples, principal label space transformation (Tai & Lin, 2010) uses the top $d$ principal components in the label space as the encoding projections:

$$\{\mathbf{v}_k\}_{k=1}^{d} \leftarrow \text{top\_d\_principal\_components}(\mathbf{Y}) \quad (12)$$

which is solved by performing SVD on the label matrix $\mathbf{Y}$ and taking the top $d$ right singular vectors. The codeword $\mathbf{z} = (\mathbf{v}_1^T \mathbf{y}, \ldots, \mathbf{v}_d^T \mathbf{y})^T$ contains the top $d$ coordinates of $\mathbf{y}$ in the principal component space.

Given predicted codeword $\hat{\mathbf{z}} = (\hat{m}_1(\mathbf{x}), \ldots, \hat{m}_d(\mathbf{x}))^T$ for a test sample $\mathbf{x}$, decoding is performed by projecting $\hat{\mathbf{z}}$ back to coordinates in the original label space and then rounding them element-wise to 0s and 1s:

$$\hat{\mathbf{y}} \leftarrow \text{round}(\mathbf{V}\hat{\mathbf{z}}) \quad (13)$$

Note that coding with principal components can potentially produce *discriminative* codewords. The top $d$ principal components provide a coordinate system that keeps as much sample variance as possible by any $d$-dimensional projections. Therefore, generated codewords for training samples tend to be spread out and far away from each other, although this does not exactly maximize the minimum codeword distance.

However, coding with principal components does not promote code predictability. Indeed, finding encoding projections as in eq. (12) is solely based on the label matrix $\mathbf{Y}$ and does not involve the input $\mathbf{X}$. As a result, this may generate codewords with large code distance but difficult to predict from the input.

### 2.4. Coding with canonical correlation analysis

Predictability for multi-label output codes is addressed in recent work (Zhang & Schneider, 2011), where output projections are obtained by canonical correlation analysis. CCA tries to find an input projection $\mathbf{u} \in R^p$ in the feature space and an output projection $\mathbf{v} \in R^q$ in the label space such that the projected variables $\mathbf{u}^T \mathbf{x}$ and $\mathbf{v}^T \mathbf{y}$ are maximally correlated:

$$\underset{\mathbf{u} \in R^p, \mathbf{v} \in R^q}{\operatorname{argmax}} \ \frac{\mathbf{u}^T \mathbf{X}^T \mathbf{Y} \mathbf{v}}{\sqrt{(\mathbf{u}^T \mathbf{X}^T \mathbf{X} \mathbf{u})(\mathbf{v}^T \mathbf{Y}^T \mathbf{Y} \mathbf{v})}} \quad (14)$$

This can be solved as a generalized eigenvalue problem, and the top $d$ pairs of eigenvectors $\{(\mathbf{u}_k, \mathbf{v}_k)\}_{k=1}^{d}$ contain the encoding projections $\{\mathbf{v}_k\}_{k=1}^{d}$.

The codeword in this method is defined as $\mathbf{z} = (y_1, \ldots, y_q, \mathbf{v}_1^T \mathbf{y}, \ldots, \mathbf{v}_d^T \mathbf{y})^T$. For a new sample $\mathbf{x}$, regression predictions for label projections are $\{\hat{m}_k(\mathbf{x})\}_{k=1}^{d}$ and classification predictions for original labels are $\{\hat{p}_{j0}(\mathbf{x}), \hat{p}_{j1}(\mathbf{x})\}_{j=1}^{q}$. Decoding is performed by maximizing a joint probability function (including $d$ Gaussian potentials from regression and $q$ Bernoulli potentials from classifiers), or equivalently minimizing the function (Zhang & Schneider, 2011):

$$\hat{\mathbf{y}} \leftarrow \underset{\mathbf{y} \in \{0,1\}^q}{\operatorname{argmin}} \ \frac{1}{2} \sum_{k=1}^{d} \frac{(\mathbf{v}_k^T \mathbf{y} - \hat{m}_k(\mathbf{x}))^2}{\hat{\sigma}_k^2}$$
$$+ \lambda \sum_{j=1}^{q} y_j \log(\frac{\hat{p}_{j0}(\mathbf{x})}{\hat{p}_{j1}(\mathbf{x})}) \quad (15)$$

where $\hat{\sigma}_k^2$ is the estimated mean squared error for regression model $\hat{m}_k$. Since the problem is defined on the label space $\mathbf{y} \in \{0,1\}^q$, approximate inference such as mean-field approximation is used for optimization.

Coding with canonical correlation analysis improves the code predictability by choosing the projection directions that are maximally correlated with the input. However, this criterion does not optimize the discriminability of the generated codewords. In other words, codewords of different outputs may be close to each other, leading to inadequate error-correcting capabilities. Consequently, even a small amount of prediction error can significantly affect the decoding result.

## 3. Maximum Margin Output Coding

In this section we propose a max-margin output coding scheme where the encoding transform promotes both discriminability and predictability of the codewords.

## 3.1. A Max-Margin Formulation

As before, codewords are predicted using regression:

$$\hat{\mathbf{M}}(\mathbf{x}) = (\hat{m}_1(\mathbf{x}), \ldots, \hat{m}_d(\mathbf{x}))^T \quad (16)$$

where each $\hat{m}_k()$ ($k = 1, \ldots, d$) is a univariate regression function for predicting $\mathbf{v}_k^T \mathbf{y}$, which is learned as in (4), and $\hat{\mathbf{M}}()$ is the corresponding multivariate regression function for the entire codeword $\mathbf{V}^T \mathbf{y}$.

For each sample $i$, the codeword $\mathbf{V}^T \mathbf{y}^{(i)}$ should be both predictable and discriminative. For predictability, we want $\hat{\mathbf{M}}(\mathbf{x}^{(i)})$ to be close to the correct codeword $\mathbf{V}^T \mathbf{y}^{(i)}$. For discriminability, we want the correct codeword $\mathbf{V}^T \mathbf{y}^{(i)}$ to have a large distance to any incorrect codeword $\mathbf{V}^T \mathbf{y}, \forall \mathbf{y} \neq \mathbf{y}^{(i)}$. In the context of prediction with output coding, it is even more strightforward and effective if the prediction $\hat{\mathbf{M}}(\mathbf{x}^{(i)})$ itself has a large distance to any incorrect codeword $\mathbf{V}^T \mathbf{y}, \forall \mathbf{y} \neq \mathbf{y}^{(i)}$.

Based on these goals, we propose the following max-margin formulation on output projections $\mathbf{V}$:

$$\operatorname*{argmin}_{\mathbf{V} \in R^{q \times d}, \{\xi_i\}_{i=1}^n} \frac{1}{2}\|\mathbf{V}\|_F^2 + \frac{C}{n}\sum_{i=1}^n \xi_i \quad (17)$$

$$\text{s.t. } \|\hat{\mathbf{M}}(\mathbf{x}^{(i)}) - \mathbf{V}^T \mathbf{y}^{(i)}\|_2^2 + \triangle(\mathbf{y}^{(i)}, \mathbf{y}) - \xi_i \quad (18)$$
$$\leq \|\hat{\mathbf{M}}(\mathbf{x}^{(i)}) - \mathbf{V}^T \mathbf{y}\|_2^2, \quad \forall \mathbf{y} \in \{0,1\}^q, \forall i$$
$$\xi_i \geq 0, \quad \forall i$$

where $\|\ \|_F$ is the Frobenius norm, $\|\ \|_2$ is the $\ell 2$ norm, $C$ is a regularization parameter, $\triangle(\mathbf{y}^{(i)}, \mathbf{y})$ is the hamming distance between binary vectors, and $\{\xi_i\}_{i=1}^n$ are slack variables, each for a training sample. With the help from slack variables, constraint (18) requires that for any sample $i$, the prediction distance to the correct codeword, denoted by $\|\hat{\mathbf{M}}(\mathbf{x}^{(i)}) - \mathbf{V}^T \mathbf{y}^{(i)}\|_2^2$, must be smaller than the prediction distance to any codeword $\|\hat{\mathbf{M}}(\mathbf{x}^{(i)}) - \mathbf{V}^T \mathbf{y}\|_2^2$ by a margin of at least $\triangle(\mathbf{y}^{(i)}, \mathbf{y})$. Note that this constraint encourages both small prediction distance to the correct codeword and large prediction distance to incorrect codewords, and hence promotes predictable and discriminative codes.

To simplify this formulation, we assume the regression functions $\hat{\mathbf{M}}(\mathbf{x}) = (\hat{m}_1(\mathbf{x}), \ldots, \hat{m}_d(\mathbf{x}))^T$ are *linear* and estimated by *least squares*. Then given training samples $(\mathbf{X}, \mathbf{Y})$, we define the $p \times q$ projection matrix $\mathbf{P}$:

$$\mathbf{P} = (\mathbf{X}^T \mathbf{X})^{-1} \mathbf{X}^T \mathbf{Y} \quad (19)$$

A small amount of regularization can be added to the diagonal of $\mathbf{X}^T \mathbf{X}$ for numerical stability. Using $\mathbf{P}$, the regression functions can be written in closed form:

$$\hat{m}_k(\mathbf{x}) = [\mathbf{P}\mathbf{v}_k]^T \mathbf{x}, \quad k = 1, 2, \ldots, d \quad (20)$$

and

$$\hat{M}(\mathbf{x}) = [\mathbf{P}\mathbf{V}]^T \mathbf{x} \quad (21)$$

Plugging eq. (21) into problem (17), we have the following max-margin formulation that is completely defined on projections $\mathbf{V}$ and slack variables $\{\xi_i\}_{i=1}^n$:

$$\operatorname*{argmin}_{\mathbf{V} \in R^{q \times d}, \{\xi_i\}_{i=1}^n} \frac{1}{2}\|\mathbf{V}\|_F^2 + \frac{C}{n}\sum_{i=1}^n \xi_i \quad (22)$$

$$\text{s.t. } \|\mathbf{V}^T(\mathbf{P}^T \mathbf{x}^{(i)} - \mathbf{y}^{(i)})\|_2^2 + \triangle(\mathbf{y}^{(i)}, \mathbf{y}) - \xi_i$$
$$\leq \|\mathbf{V}^T(\mathbf{P}^T \mathbf{x}^{(i)} - \mathbf{y})\|_2^2, \quad \forall \mathbf{y} \in \{0,1\}^q, \forall i$$
$$\xi_i \geq 0, \quad \forall i$$

## 3.2. Metric Learning Formulation

Problem (22) is a quadratic program with quadratic constraints, and we first convert it to a metric learning problem. Define $q \times q$ matrix $\mathbf{Q}$:

$$\mathbf{Q} = \mathbf{V}\mathbf{V}^T \quad (23)$$

which is the Mahalanobis distance metric induced by $\mathbf{V}$. Also, define a set of new feature vectors:

$$\phi_{i\mathbf{y}} = \mathbf{P}^T \mathbf{x}^{(i)} - \mathbf{y}, \quad \forall \mathbf{y} \in \{0,1\}^q, \forall i \quad (24)$$

Now we formulate the metric learning problem as:

$$\operatorname*{argmin}_{\mathbf{Q} \in S_q^+, \{\xi_i\}_{i=1}^n} \frac{1}{2}\operatorname{trace}(\mathbf{Q}) + \frac{C}{n}\sum_{i=1}^n \xi_i \quad (25)$$

$$\text{s.t. } \phi_{i\mathbf{y}^{(i)}}^T \mathbf{Q} \phi_{i\mathbf{y}^{(i)}} + \triangle(\mathbf{y}^{(i)}, \mathbf{y}) - \xi_i$$
$$\leq \phi_{i\mathbf{y}}^T \mathbf{Q} \phi_{i\mathbf{y}}, \quad \forall \mathbf{y} \in \{0,1\}^q, \forall i$$
$$\xi_i \geq 0, \quad \forall i$$

where $\mathbf{Q} \in S_q^+$ is positive semidefinite. The objective function and constraints are linear in $\mathbf{Q}$ and $\{\xi_i\}_{i=1}^n$.

We briefly show the equivalence between problem (22) and (25) as follows. For any feasible solution $\mathbf{V}$ to (22), we can define $\mathbf{Q} = \mathbf{V}\mathbf{V}^T \in S_q^+$. Also, for any feasible solution $\mathbf{Q}$ to (25), since $\mathbf{Q}$ is positive semidefinite and thus has no negative eigenvalue, we can define $\mathbf{V}$ as:

$$\mathbf{V} = \mathbf{Q}^{\frac{1}{2}} = \mathbf{U}\mathbf{D}^{\frac{1}{2}} \quad (26)$$

where the $q \times q$ matrix $\mathbf{U}$ contains (as columns) the $q$ eigenvectors of $\mathbf{Q}$, and $\mathbf{D}$ is the diagonal matrix of eigenvalues. Given this one-to-one mapping between $\mathbf{V}$ and $\mathbf{Q}$, we have $\operatorname{trace}(\mathbf{Q}) = \|\mathbf{V}\|_F^2$ and $\phi_{i\mathbf{y}}^T \mathbf{Q} \phi_{i\mathbf{y}} = \|\mathbf{V}^T(\mathbf{P}^T \mathbf{x}^{(i)} - \mathbf{y})\|_2^2$. Therefore, any feasible (or optimal) solution to (25) gives a feasible (or optimal) solution to (22), and vice versa.

### 3.3. Incorporating Original Labels and Their Classifiers

As shown in eq. (3), the codeword can also include $q$ original labels, i.e., $\mathbf{z} = (y_1, \ldots, y_q, \mathbf{v}_1^T\mathbf{y}, \ldots, \mathbf{v}_d^T\mathbf{y})^T$. Classifiers $\{\hat{p}_j\}_{j=1}^q$ can be learned to predict original labels as in (5), and the decoding algorithm can make use of both regression and classifier outputs, e.g., as in eq. (15). In this case, the encoding projection should also be aware of the original labels $(y_1, \ldots, y_q)$ in the codeword, so that the projection part $(\mathbf{v}_1^T\mathbf{y}, \ldots, \mathbf{v}_d^T\mathbf{y})$ can provide *complementary* information.

To adapt our max-margin formulation (25) to this new information, we assume that classifiers $\{\hat{p}_j\}_{j=1}^q$ have already been learned, and thus for each sample $\mathbf{x}$ we know the classifier output $\hat{p}_{j0}(\mathbf{x}) = P(y_j = 0|\mathbf{x})$ and $\hat{p}_{j1}(\mathbf{x}) = P(y_j = 1|\mathbf{x})$. We have the new formulation:

$$\operatorname*{argmin}_{\mathbf{Q}\in S_q^+, \{\xi_i\}_{i=1}^n} \frac{1}{2}\operatorname{trace}(\mathbf{Q}) + \frac{C}{n}\sum_{i=1}^n \xi_i \qquad (27)$$

$$\text{s.t.} \quad \phi_{i\mathbf{y}^{(i)}}^T \mathbf{Q}\phi_{i\mathbf{y}^{(i)}} - \log P(\mathbf{y}^{(i)}|\mathbf{x}^{(i)}) + \triangle(\mathbf{y}^{(i)}, \mathbf{y}) - \xi_i$$
$$\leq \phi_{i\mathbf{y}}^T \mathbf{Q}\phi_{i\mathbf{y}} - \log P(\mathbf{y}|\mathbf{x}^{(i)}), \quad \forall \mathbf{y} \in \{0,1\}^q, \forall i$$
$$\xi_i \geq 0, \quad \forall i$$

where

$$P(\mathbf{y}|\mathbf{x}^{(i)}) = \prod_{j=1}^q P(y_j|\mathbf{x}^{(i)}) = \prod_{j=1}^q \hat{p}_{jy_j}(\mathbf{x}^{(i)}) \qquad (28)$$

is the joint probability of label vector $\mathbf{y} = (y_1, \ldots, y_q)$ on sample $\mathbf{x}^{(i)}$ predicted by classifiers $\{\hat{p}_j\}_{j=1}^q$.

In this new formation (27), $\phi_{i\mathbf{y}}^T\mathbf{Q}\phi_{i\mathbf{y}}$ is extended into $\phi_{i\mathbf{y}}^T\mathbf{Q}\phi_{i\mathbf{y}} - \log P(\mathbf{y}|\mathbf{x}^{(i)})$. Recall that $\phi_{i\mathbf{y}}^T\mathbf{Q}\phi_{i\mathbf{y}}$ is equivalent to $\|\mathbf{V}^T(\mathbf{P}^T\mathbf{x}^{(i)} - \mathbf{y})\|_2^2$ in (22), which is the distance between the regression prediction on the $i$th sample and the encoding of the label vector $\mathbf{y}$. We expect that the correct label vector $\mathbf{y}^{(i)}$ should lead to lower values on this term than other $\mathbf{y}$. Similarly, $\log P(\mathbf{y}|\mathbf{x}^{(i)})$ is the log-probability of $\mathbf{y}$ predicted by classifiers on sample $i$, and we expect that $\mathbf{y}^{(i)}$ should give higher values on this term than other label vectors $\mathbf{y}$. As a result, we now use the combined term $\phi_{i\mathbf{y}}^T\mathbf{Q}\phi_{i\mathbf{y}} - \log P(\mathbf{y}|\mathbf{x}^{(i)})$ to measure the margin between correct and incorrect outputs. The main outcome of this new formulation is that distance metric $\mathbf{Q}$ will focus on the constraints where $-\log P(\mathbf{y}|\mathbf{x}^{(i)})$ alone is not strong enough to ensure the margin. In other words, the output coding concentrates on the cases where classifiers $\{\hat{p}_j\}_{j=1}^q$ alone tend to misclassify.

### 3.4. Cutting Plane Method with Overgenerating

In this section we consider how to solve problem (27). This problem involves an exponentially large number of constraints due to the combinatorial nature of the label space $\{0,1\}^q$. As studied in structured prediction (Tsochantaridis et al., 2004; Taskar et al., 2003), problem (27) could be solved efficiently, e.g., by the cutting-plane method, if a computationally tractable separate oracle exists to determine which of the exponentially many constraints is most violated (Tsochantaridis et al., 2004). However, without a specific structure (e.g., a chain or a tree) in the label space to enable efficient inference, the separate oracle for problem (27) is computationally intractable.

To address this issue, we use overgenerating (i.e., relaxation) (Finley & Joachims, 2008) with the cutting plane method. To use overgenerating technique, we need to relax $\forall \mathbf{y} \in \{0,1\}^q$ in the constraint of (27) to a continuous domain, e.g., $\forall \mathbf{y} \in [0,1]^q$. However, $\triangle(\mathbf{y}^{(i)}, \mathbf{y})$ and $\log P(\mathbf{y}|\mathbf{x}^{(i)})$ in (27) are only defined on $\mathbf{y} \in \{0,1\}^q$. To handle this, we redefine $\triangle(\mathbf{y}^{(i)}, \mathbf{y})$ as

$$\widetilde{\triangle}(\mathbf{y}^{(i)}, \mathbf{y}) = \|\mathbf{y}^{(i)} - \mathbf{y}\|_1 = \sum_{j=1}^q |y_j^{(i)} - y_j| \qquad (29)$$

Then noticing $\log P(\mathbf{y}|\mathbf{x}^{(i)}) = \sum_{j=1}^q \log P(y_j|\mathbf{x}^{(i)})$, we redefine:

$$\log\widetilde{P}(\mathbf{y}|\mathbf{x}^{(i)}) = \sum_{j=1}^q \log\widetilde{P}(y_j|\mathbf{x}^{(i)}) \qquad (30)$$

where each $\log\widetilde{P}(y_j|\mathbf{x}^{(i)})$ is the linear interpolation of $\log P(y_j = 0|\mathbf{x}^{(i)})$ and $\log P(y_j = 1|\mathbf{x}^{(i)})$.

Using (29) and (30), the new relaxed problem is:

$$\operatorname*{argmin}_{\mathbf{Q}\in S_q^+, \{\xi_i\}_{i=1}^n} \frac{1}{2}\operatorname{trace}(\mathbf{Q}) + \frac{C}{n}\sum_{i=1}^n \xi_i \qquad (31)$$

$$\text{s.t.} \quad \phi_{i\mathbf{y}^{(i)}}^T \mathbf{Q}\phi_{i\mathbf{y}^{(i)}} - \log\widetilde{P}(\mathbf{y}^{(i)}|\mathbf{x}^{(i)}) + \widetilde{\triangle}(\mathbf{y}^{(i)}, \mathbf{y}) - \xi_i$$
$$\leq \phi_{i\mathbf{y}}^T \mathbf{Q}\phi_{i\mathbf{y}} - \log\widetilde{P}(\mathbf{y}|\mathbf{x}^{(i)}), \quad \forall \mathbf{y} \in [0,1]^q, \forall i$$
$$\xi_i \geq 0, \quad \forall i$$

where $\forall \mathbf{y} \in \{0,1\}^q$ in (27) is relaxed to $\forall \mathbf{y} \in [0,1]^q$.

This new problem can be solved by the cutting plane method, because the separate oracle (i.e., finding the most violated constraint for each sample $i$) is:

$$\operatorname*{argmin}_{\mathbf{y}\in[0,1]^q} \phi_{i\mathbf{y}}^T \mathbf{Q}\phi_{i\mathbf{y}} - \log\widetilde{P}(\mathbf{y}|\mathbf{x}^{(i)}) - \widetilde{\triangle}(\mathbf{y}^{(i)}, \mathbf{y}) \qquad (32)$$

where $\log\widetilde{P}(\mathbf{y}|\mathbf{x}^{(i)})$ and $\widetilde{\triangle}(\mathbf{y}^{(i)}, \mathbf{y})$ are linear in $\mathbf{y}$, and $\phi_{i\mathbf{y}}^T\mathbf{Q}\phi_{i\mathbf{y}}$ is quadratic in $\mathbf{y}$ given $\phi_{i\mathbf{y}}$ defined as (24). So (32) is a simple box-constrained quadratic program.

Maximum Margin Output Coding

### 3.5. Encoding and Decoding

After solving **Q** in (31), encoding projections are obtained as (26), and one can choose $d$, the number of projections, by keeping only the first $d$ columns of **V** in (26) for any $d \leq q$. The codeword as in (3) includes original labels, and decoding is performed as (15).

## 4. Empirical Study

**Data**. We perform experiments on three real-world data sets[1]: an image data set (*Scene*), a text data set (*Medical*) and a music data set (*Emotions*). *Scene* is an image collection for outdoor scene recognition. Each image is represented by 294 dimensional color features and labeled as: beach, sunset, fall foliage, field, mountain and urban. *Emotions* is a music classification problem. Each song is represented by 72 rhythmic and timbre features, and tagged with six emotions: amazed, happy, relaxed, quiet, sad and angry. *Medical* is a clinical text collection, where each document is represented by 1449 words and labeled with ICD-9-CM codes. Many labels in *Medical* are rare, so we select the 10 most common labels to study.

**Methods**. We compare the proposed max-margin output coding scheme to several recently proposed multi-label output codes as well as a number of other multi-label classification methods:

- *Binary relevance (BR)*. This baseline method learns to classify each label independently. It is also called one-vs-all decomposition.
- *Coding with compressed sensing (CodingCS)* (Hsu et al., 2009). As reviewed in Section 2.2, this method uses random projections for encoding and sparse approximation for decoding. Specifically, we use CoSaMP (Needell & Tropp, 2008) for decoding.
- *Coding with PCA (CodingPCA)* (Tai & Lin, 2010). As reviewed in Section 2.3, this method uses principal components for encoding, and PCA reconstruction and rounding for decoding.
- *Coding with PCA-Redundant (CodingPCA-R)*. CodingPCA does not include original labels into the codeword. We also try this option to produce more redundancy as in eq. (3). Decoding follows eq. (15).
- *Coding with CCA (CodingCCA)* (Zhang & Schneider, 2011). As reviewed in Section 2.4, this method uses CCA for encoding. Decoding follows eq. (15).
- *Calibrated label ranking (CLR)* (Fürnkranz et al., 2008). This method combines both one-vs-one and one-vs-all classifiers for multi-label classification. It can also be considered as an output coding method.

---
[1] http://mulan.sourceforge.net/

*Table 1.* Subset accuracy on Scene data set: mean and standard error over 30 random runs

| Method (#Base Models) | Mean | Standard Error |
|---|---|---|
| $BR(6)$ | 0.4238 | 0.0040 |
| $CodingCS(100)$ | 0.3821 | 0.0047 |
| $CodingPCA(6)$ | 0.3691 | 0.0075 |
| $CodingPCA\text{-}R(12)$ | 0.4305 | 0.0046 |
| $CodingCCA(12)$ | 0.4928 | 0.0090 |
| $CLR(21)$ | 0.4218 | 0.0034 |
| $LEAD(6)$ | 0.4547 | 0.0048 |
| $MaxMargin(12)$ | **0.5448** | 0.0073 |

*Table 2.* Macro-F1 score on Scene data set: mean and standard error over 30 random runs

| Method (#Base Models) | Mean | Standard Error |
|---|---|---|
| $BR(6)$ | 0.6209 | 0.0029 |
| $CodingCS(100)$ | 0.5234 | 0.0050 |
| $CodingPCA(6)$ | 0.5279 | 0.0096 |
| $CodingPCA\text{-}R(12)$ | 0.6049 | 0.0039 |
| $CodingCCA(12)$ | 0.6312 | 0.0038 |
| $CLR(21)$ | 0.6238 | 0.0026 |
| $LEAD(6)$ | 0.5958 | 0.0042 |
| $MaxMargin(12)$ | **0.6462** | 0.0046 |

*Table 3.* Micro-F1 score on Scene data set: mean and standard error over 30 random runs

| Method (#Base Models) | Mean | Standard Error |
|---|---|---|
| $BR(6)$ | 0.6117 | 0.0030 |
| $CodingCS(100)$ | 0.5345 | 0.0041 |
| $CodingPCA(6)$ | 0.5404 | 0.0076 |
| $CodingPCA\text{-}R(12)$ | 0.6014 | 0.0032 |
| $CodingCCA(12)$ | 0.6251 | 0.0035 |
| $CLR(21)$ | 0.6163 | 0.0024 |
| $LEAD(6)$ | 0.6002 | 0.0036 |
| $MaxMargin(12)$ | **0.6382** | 0.0047 |

- *Multi-label learning by exploiting label dependency (LEAD)* (Zhang & Zhang, 2010). This method learns a Bayes network on labels and use it to capture label dependency in multi-label classification.
- *Max-Margin coding (MaxMargin)*. Our max-margin coding formulation where encoding is obtained by solving (31) and (26). Decoding follows eq. (15).

**Evaluation measures**. We consider three evaluation measures for multi-label classification:

- Subset accuracy: rates of correctly classifying *all* the labels. It is difficult to achieve high subset accuracy.
- Macro-averaged F-1 score: calculate the F-1 score for each label and take the average over labels. F-1

Maximum Margin Output Coding

Table 4. Subset accuracy on Medical data set: mean and standard error over 30 random runs

| Method (#Base Models) | Mean | Standard Error |
|---|---|---|
| $BR(10)$ | 0.7673 | 0.0039 |
| $CodingCS(100)$ | 0.7071 | 0.0019 |
| $CodingPCA(10)$ | 0.7541 | 0.0041 |
| $CodingPCA\text{-}R(20)$ | 0.7803 | 0.0031 |
| $CodingCCA(20)$ | 0.7824 | 0.0029 |
| $CLR(55)$ | 0.7632 | 0.0037 |
| $LEAD(10)$ | 0.7718 | 0.0038 |
| $MaxMargin(20)$ | **0.7930** | 0.0042 |

Table 7. Subset accuracy on Emotions data set: mean and standard error over 30 random runs

| Method (#Base Models) | Mean | Standard Error |
|---|---|---|
| $BR(6)$ | 0.2264 | 0.0031 |
| $CodingCS(100)$ | 0.1665 | 0.0035 |
| $CodingPCA(6)$ | 0.2198 | 0.0027 |
| $CodingPCA\text{-}R(12)$ | 0.2601 | 0.0024 |
| $CodingCCA(12)$ | 0.3005 | 0.0040 |
| $CLR(21)$ | 0.2266 | 0.0037 |
| $LEAD(6)$ | 0.1559 | 0.0035 |
| $MaxMargin(12)$ | **0.3114** | 0.0042 |

Table 5. Macro-F1 score on Medical data set: mean and standard error over 30 random runs

| Method (#Base Models) | Mean | Standard Error |
|---|---|---|
| $BR(10)$ | 0.8626 | 0.0029 |
| $CodingCS(100)$ | 0.7987 | 0.0026 |
| $CodingPCA(10)$ | 0.8523 | 0.0032 |
| $CodingPCA\text{-}R(20)$ | 0.8697 | 0.0021 |
| $CodingCCA(20)$ | 0.8703 | 0.0020 |
| $CLR(55)$ | 0.8556 | 0.0029 |
| $LEAD(10)$ | 0.8550 | 0.0035 |
| $MaxMargin(20)$ | **0.8710** | 0.0039 |

Table 8. Macro-F1 score on Emotions data set: mean and standard error over 30 random runs

| Method (#Base Models) | Mean | Standard Error |
|---|---|---|
| $BR(6)$ | 0.6248 | 0.0027 |
| $CodingCS(100)$ | 0.4742 | 0.0043 |
| $CodingPCA(6)$ | 0.5592 | 0.0041 |
| $CodingPCA\text{-}R(12)$ | 0.6367 | 0.0028 |
| $CodingCCA(12)$ | 0.6539 | 0.0027 |
| $CLR(21)$ | 0.6197 | 0.0029 |
| $LEAD(6)$ | 0.4512 | 0.0046 |
| $MaxMargin(12)$ | **0.6609** | 0.0029 |

Table 6. Micro-F1 score on Medical data set: mean and standard error over 30 random runs

| Method (#Base Models) | Mean | Standard Error |
|---|---|---|
| $BR(10)$ | 0.8785 | 0.0022 |
| $CodingCS(100)$ | 0.8333 | 0.0013 |
| $CodingPCA(10)$ | 0.8780 | 0.0020 |
| $CodingPCA\text{-}R(20)$ | 0.8853 | 0.0018 |
| $CodingCCA(20)$ | 0.8867 | 0.0018 |
| $CLR(55)$ | 0.8757 | 0.0022 |
| $LEAD(10)$ | 0.8736 | 0.0023 |
| $MaxMargin(20)$ | **0.8919** | 0.0025 |

Table 9. Micro-F1 score on Emotions data set: mean and standard error over 30 random runs

| Method (#Base Models) | Mean | Standard Error |
|---|---|---|
| $BR(6)$ | 0.6363 | 0.0024 |
| $CodingCS(100)$ | 0.5260 | 0.0029 |
| $CodingPCA(6)$ | 0.5957 | 0.0032 |
| $CodingPCA\text{-}R(12)$ | 0.6482 | 0.0026 |
| $CodingCCA(12)$ | 0.6633 | 0.0025 |
| $CLR(21)$ | 0.6331 | 0.0025 |
| $LEAD(6)$ | 0.5142 | 0.0038 |
| $MaxMargin(12)$ | **0.6688** | 0.0027 |

score is popular since the distribution of positives and negatives for a label is usually imbalanced.

- Micro-averaged F-1 score: aggregate true positives, true negatives, false positives and false negatives over labels, and then calculate an overall F-1 score.

**Experimental settings.** On each data set, we perform 30 random runs and report means and standard errors of each evaluation measure. The number of training samples in each random run is set to 300.

For *CodingCS*, the number of projections $d$ is set to 100 to provide highly redundant codewords. For *CodingPCA*, *CodingPCA-R*, *CodingCCA* and *MaxMargin*, the number of output projections is set to the maximum possible number: the number of original labels.

For all methods, base regression models are ridge regression and base classifiers are $\ell 2$-penalized logistic regression, and their regularization parameters are chosen by cross validation. For LEAD, the Bayes net is learned using the score-based searching algorithm in the Bayesian Net Toolbox[2]. For decoding that follows (15), $\lambda$ is set to 1, i.e., classifiers and regression models are equally weighted in decoding. The parameter $C$ in (31) is set to $10^6$. Most methods need to round their final predictions into 0/1 assignments (e.g., from a probability forecast or a relaxed solution to the la-bel assignment), and in these cases we use 0.5 as the threshold without further optimization.

---
[2]http://code.google.com/p/bnt/

**Empirical Results.** Results for the Scene data set are shown in Table 1 - Table 3; results for Medical are shown in Table 4 - Table 6; results for Emotions are shown in Table 7 - Table 9. Each table contains one evaluation measure. From the results we can see:

- BR provides a solid baseline with good performance.
- CodingCS generally underperforms, indicating that encoding with random projections is not effective.
- CodingPCA-R outperforms CodingPCA because CodingPCA-R uses more redundant codewords.
- LEAD's performance is not stable across data sets. Structure learning for bayes nets is still challenging.
- CLR performs comparably to BR, despite the fact that it is one of the most redundant methods in terms of the number of base models used.
- CodingPCA-R, CodingCCA and MaxMargin are most successful. Their codewords include both label projections and original labels, and their decodings combine both regression and classification outputs.
- MaxMargin outperforms CodingCCA and CodingPCA, because max-margin encoding promotes both code discriminability and code predictability.
- CodingCCA performs better than CodingPCA-R, showing the importance of predictable codewords.

## 5. Related Work

Our work follows the direction of multi-label output coding (Hsu et al., 2009) and is motivated by the recent success of coding with PCA (Tai & Lin, 2010) and CCA (Zhang & Schneider, 2011) and their connections to code distance and code predictability. Our max-margin formulation is converted into a metric learning problem, as in (Weinberger et al., 2006), but with a metric defined for the label space and an exponential number of constraints caused by label combinations. The optimization technique developed for structured prediction (Tsochantaridis et al., 2004; Taskar et al., 2003), more specically the cutting plane method with overgenerating (Finley & Joachims, 2008), is used to solve our metric learning problem.

## 6. Conclusion

Discriminability and predictability are both important for output codes. In this paper we propose a max-margin formulation for multi-label output coding, which promotes both discriminative and predictable codes. We convert this formulation into a metric learning problem in the label space, and combine overgener-ating with the cutting plane method for optimization. Our method outperforms many existing methods on multi-label image, text and music data sets.